\documentclass{ieeeaccess}
\usepackage{cite}
\usepackage{amsmath,amssymb,amsfonts}
\usepackage{algorithmic}
\usepackage{graphicx}
\usepackage{textcomp}
\usepackage{subfigure}

\def\BibTeX{{\rm B\kern-.05em{\sc i\kern-.025em b}\kern-.08em
    T\kern-.1667em\lower.7ex\hbox{E}\kern-.125emX}}
\begin{document}
\history{Date of publication xxxx 00, 0000, date of current version xxxx 00, 0000.}
\doi{10.1109/ACCESS.2023.0322000}

\title{Algorithmic decision making methods for fair credit scoring}
\author{\uppercase{Darie Moldovan}\authorrefmark{1,2}, \IEEEmembership{Member, IEEE},
}

\address[1]{Faculty of Economics and Business Administration, Babe\c{s}-Bolyai Univesity, Cluj-Napoca, Romania}
\address[2]{Faculty of Economics and Business Administration, Alexandru Ioan Cuza University, Ia\c{s}i, Romania}

\tfootnote{This work was supported in part by the 2022 Development Fund of the UBB. It was also partly supported by the Romanian National Authority for Scientific Research Grant no. PN-III-P4-ID-PCE-2020-0929.}

\markboth
{D. Moldovan: Algorithmic decision making methods for fair credit scoring}
{D. Moldovan: Algorithmic decision making methods for fair credit scoring}

\corresp{Corresponding author: Darie Moldovan (e-mail: darie.moldovan@ubbcluj.ro).}

\begin{abstract}
The effectiveness of machine learning in evaluating the creditworthiness of loan applicants has been demonstrated for a long time. However, there is concern that the use of automated decision-making processes may result in unequal treatment of groups or individuals, potentially leading to discriminatory outcomes. This paper seeks to address this issue by evaluating the effectiveness of 12 leading bias mitigation methods across 5 different fairness metrics, as well as assessing their accuracy and potential profitability for financial institutions. Through our analysis, we have identified the challenges associated with achieving fairness while maintaining accuracy and profitabiliy, and have highlighted both the most successful and least successful mitigation methods. Ultimately, our research serves to bridge the gap between experimental machine learning and its practical applications in the finance industry.
\end{abstract}


\titlepgskip=-21pt

\maketitle

\section{Introduction}\label{sec1}

Credit scoring applications play an important role in modern society, and the approval process of loans increasingly migrates from human decisions to complex algorithmic decisions.  Agarwal et al. \cite{agarwal2021ai} discussed the important benefits of an automated decision process for financial institutions, such as growing the business while lowering costs, increasing approval rates without increasing credit risks, and providing a more seamless application process for the client.

This business reality raises questions regarding the supervision of such automated decisions. Blattner et al. \cite{blattner2021unpacking} discussed how algorithmic governance faces a trade-off between complexity (read performance) and oversight (read: capacity to audit). The interpretability of the models, particularly the credit scoring methods, has long been a source of concern for industry regulators (e.g., national banks, financial authorities, and governments). Their interest was related to the capability of a human to understand how the decision was taken in order to supervise, mitigate risks, and prevent the misconduct of financial institutions. An automatic decision, even if interpretable, may lead to different treatments for groups or individuals, defined by some specific attributes, eventually causing discrimination \cite{hurlin2021fairness}. 

While the laws explicitly cover some discrimination factors (such as gender, race, and nationality), other unrestricted information may be used to discriminate against vulnerable groups (e.g., based on behavior and wealth). Moreover, information from external public sources such as social networks may be used as correlated information, potentially leading to traits of race or gender \cite{neverending}.

There is a certain concern (among researchers, but also at the government level) about the use of AI in decision making, and several reports  \cite{european2020white, un2021} indicate future regulatory guidelines for this field. Both interpretability and fairness are addressed, together with privacy (already regulated by the GDPR in the EU\footnote{The General Data Protection Regulation}), technical robustness and accountability (with specific requirements in force through Basel III\footnote{An internationally agreed regulatory framework developed by the Basel Committee on Banking Supervision in response to the financial crisis of 2007-09.} and IFRS9 \footnote{Accounting regulations, requiring financial institutions to implement forward-looking estimates with respect to expected losses in their financial statements.} regulations). 

Our study addresses the problem of ensuring fairness in the automatic credit scoring process using machine learning algorithms. Since the research interest for ensuring fair processing in automatic decision-making started to grow, several processing methods and evalution metrics were defined. As we will show, there are serious difficulties in implementing these fairness processors in a real-world context, causing a gap between lab development and industrialized implementation. We therefore benchmark state-of-the-art methods in this field by considering different fairness definitions from the literature. Our empirical findings were obtained by applying these methods to two datasets containing retail credit applications, one of which is a novel real-world dataset relating to the Romanian consumer loan market.

While the financial industry primarily focuses on financial risk assessments when evaluating loan applications, factors such as equal opportunity are either imposed by regulations or considered indirectly in the loan granting process. However, experiments in the literature have shown that discrimination is present in this type of process. Therefore, a framework for benchmarking different bias mitigation techniques could prove useful for systematically analyzing and addressing potential biases. This framework would allow individuals or groups who might have historically been underserved or disadvantaged to access credit on fair terms. Additionally, implementing a fair credit scoring framework would demonstrate a commitment to transparency, fairness, and ethical lending practices by financial institutions.

The main contributions of our work are as follows:
\begin{enumerate}
  \item We help bridge the gap between lab development and industrialized machine learning by identifying the strengths and weaknesses of 12 bias mitigation methods with the help of a real-world data environment implementation, which is currently the most inclusive comparison of this type.
  \item We show empirical evidence from a credit scoring case study indicating how bias mitigation methods perform from three different perspectives: the amelioration of fairness metrics, balanced accuracy of the models, and expected profit for financial institutions.
  \item This study makes a new dataset in the field of credit scoring available to the public, allowing for the replication of the results and the development of novel experiments in this field where real data is scarce.
  \item Finally, we review in depth the fairness metrics found in the machine-learning fairness literature, together with the most prominent fairness processors that have emerged in the last decade.
\end{enumerate}

The remainder of this study is organized as follows. The literature review section describes the state of the art in the area of fairness in machine learning, with a special section dedicated to credit-scoring fairness developments. The methodology describes the general framework for applying fairness processors to data based on three routes found in the literature: pre-processing , in-processing, and post-processing . This section describes the fairness metrics used in the experiments. Next, the study presents the experimental setup, including data pre-processing. The Results section presents and discusses the outcomes of the case study. The last section concludes the study and provides insights for future research.

\bigskip

\section{Related work}\label{sec2}

\subsection{Fairness in machine learning}\label{subsec21}
Research interest in the fairness of machine learning has increased over the last decade, with several fairness measures arising in the literature.  As Barocas et al. \cite{barocas-hardt-narayanan} note, different measures are based on various intuitions regarding fairness. The following represent examples of the numerous attempts of defining a comprehensive fairness metric: individual fairness - advocates for similar treatment of similar cases \cite{dworkfairness}; statistical parity - considers an individual belonging to a protected group to bear the same risk score as all the other members \cite{hardt2016equality};  accuracy fairness (or accuracy equity) - implies different treatments for different cases, assuming a perfect classifier is also fair\cite{dieterich2016compas}; threshold fairness - considers the same decision threshold should be applied for each group (protected or unprotected) \cite{simoiu2017problem};  and calibration - conditions the estimates to have the same effectiveness for all individuals \cite{kleinberg2017inherent}.  The majority of viewpoints concern equal treatment in general across different groups, as defined by protected attributes (also known as sensitive attributes), such as gender, nationality, and race (statistical parity), or, more specifically, equal treatment based on some constraints, such as predictive parity and threshold fairness. 

Generally, the literature sources agree that not all fairness criteria can be used simultaneously to evaluate the fairness of an ML process \cite{paassen2019dynamic,bono2021algorithmic, parkes2019algorithmic, kleinberg2017inherent},  as some of them  contradict others. As a result, while we review all of the criteria suggested by the literature in this field from a credit scoring standpoint, we do not expect to find a method that can satisfy all fairness criteria simultaneously.

Grouping the fairness metrics, three fairness criteria have emerged as standards for measuring group fairness: separation, independence, and sufficiency \cite{barocas-hardt-narayanan, castelnovo2022clarification}.

The \textit{separation} criterion, also known as \textit{equality of odds}, evaluates the classification in such way that false positive rate (e.g.,  FP - clients wrongly associated with a high risk of not repaying debt) equal the false negative rate (e.g., FN - clients wrongly associated with a low risk of not repaying debt) for each subgroup (protected and unproctected), this being guaranteed for any cutoff used. Kozodoi et al. \cite{kozodoi2022fairness} suggested a method for measuring separation as the average absolute difference between the group-wise FP and FN rates (FPR and FNR):
\begin{equation}
\begin{split}
SP &= \frac{1}{2}\lvert(FPR_{(D=unprivileged)}-FPR_{(D=privileged})+\\ &+(FNR_{(D=unprivileged)}-FNR_{(D=privileged)})\rvert\label{eq1} 
\end{split}
\end{equation} 
where $x_a$ represents protected attributes. In this way, a value of SP$\approx$0 suggests perfect separation, while high values of SP suggest stronger discrimination, but without showing which group is privileged.
From a slightly different perspective, Hardt et al. \cite{hardt2016equality} initially proposed a measure where the difference between FP and FN rates is considered, also targeting a value close to 0 for a fair classification.

The \textit{independence} criterion (also found in the literature as \textit{demographic parity and statistical parity}) concerns the model prediction of possible dependency on the protected attribute. In other words, credit scores should be distributed in the same way within each subgroup if the predictions are not affected by the protected attribute. Barocas et al. \cite{barocas-hardt-narayanan} measured the connection between the protected attribute and the score (mutual information) using entropy ($H$):
\begin{align}
I(x_a;R)=H(x_a)+H(R)-H(x_a,R)\label{eq2}
\end{align}
where $R$ is the score. The independence criterion would be satisfied if $I(x_a;R)$$\approx$0.

However, this can be difficult to achieve in practice in the field of credit scoring because some of the protected attributes (e.g. race, gender) are linked to significant differences in wealth and income \cite{bono2021algorithmic}. By strictly applying this criterion, the scorecard would induce a higher risk appraisal for some groups while reducing vigilance for other groups. This can lead to a significant distortion of the reality due to an increase in FN and FP, ultimately being harmful for those not truly affording the indebtedness, denying access to loans to those under-appreciated and, on the sell side, making the lending business unpractical. Several studies have defined a relaxed measure of independence, attempting to make it usable \cite{barocas-hardt-narayanan, kozodoi2022fairness}. Instead of trying to reduce $I(x_a;R)$ to 0, the ratio $\epsilon$ could be considered acceptable, with an indication of 0.2 as a reference value \cite{feldman}. 

\textit{Sufficiency} implies that the score already includes  protected attributes when predicting the target variable \cite{barocas-hardt-narayanan}. This means that clients with the same credit score carry the same risk of not reimbursing the loan, regardless of whether they belong to a protected group. This interpretation is closely related to the notion of \textit{callibration by group}, which is formally defined as:
\begin{align}
Pr\{\textit{Y}=1\mid R=r, D=d\}=r\label{eq3}
\end{align}
for all scores \textit{r} and groups \textit{d}. For this approach to be valid, the score must be considered as a probability. \cite{barocas-hardt-narayanan}note that, based on practical experiments, sufficiency is already satisfied to some degree owing to machine learning inner mechanisms, and trying to impose sufficiency when training a model will result in only trivial improvements. 

Another trend in the AI fairness literature is the observation of  \textit{individual fairness} in  data, which attempts to offset the limitations of the group fairness assessment methods discussed above. One problem with group fairness is the propagation of an error (e.g. from a small group), leading to discrimination in subgroups, even if group fairness can still be achieved \cite{pmlr-v80-kearns18a}. Focusing on statistical fairness leads to the impossibility of achieving the desired protection for individuals. In other words, the fairness achieved by considering individual attributes will not translate into achieving fairness for the combination of two or more of those attributes.  Speicher et al. \cite{speicher2018unified} employed economic inequality indices to measure the outcomes of an algorithm towards individuals or groups. Their findings also suggest that eliminating the unfair treatment between groups may lead to an increase in unfairness within a group (at the individual level).

While the fairness of algorithmic decision-making has become a hot topic for debate and research, we noticed credit scoring as one of the most popular examples in the literature when referring to possible discrimination caused by machine learning methods.

\subsection{Credit scoring application fairness}\label{subsubsec22}

The research community studying modern machine learning techniques for credit scoring has recently turned their attention towards the implications of the distributional impact of algorithmic decisions on sensitive/vulnerable groups. While traditional methods for credit scoring (e.g., human application analysis) were outpaced by machine learning techniques in terms of profitability for financial institutions, evidence shows how the latter produces predictions with greater variance \cite{fuster2022predictably}.

Zarsky \cite{zarsky2016trouble} attempted to create an analytic framework for an orderly debate on the subject of algorithmic decision-making, focusing on credit scoring to substantiate the proposal. Two dimensions are considered the main triggers for concerns: the problems generated by machine learning reasoning and the attributes that exacerbate these problems. The two problems identified as debatable are the efficiency (or inefficiency) of the automated decision making and the second is, obviously, the fairness of the process. One attribute involved in the debate is the automation of the process, which can show solid results at an aggregated level but it is still prone to errors because of inaccuracies in the training data or specific algorithm settings (also identified in \cite{hurley2016credit} to be a possible burden for the consumer). The alternative human approach is also subject to biased decision making. Transparency is considered a solution to improve automated processing, by providing the opportunity to improve/correct the data, but may also lead to increasing costs.

Kozodoi et al. \cite{kozodoi2022fairness}  assess the feasibility of a fair credit scoring system in a profit-driven environment. They deal with the strong inverse proportionality between fairness and profitability and show how the possibility of reducing discrimination can be achieved at very reasonable costs for financial institutions. However, a completely fair scoring system appears to be a pipe dream, as profitability is stifled by overly strict conditions and an increasing risk of default.

Another question addressed by the fair credit scoring literature challenges the efficiency of imposing protection for specific groups. Liu et al. \cite{liu2018delayed} attempted to quantify the impact of fairness constrained classifications over time . They assume that the lender will always try to maximize the utility of the models (profit) irrespective of the fairness constraints. In other words, applying fairness criteria could have unwanted effects over time, both for a protected group (a defaulted client will worsen their credit worthiness) and for financial institutions (credit defaults considerably impact the profitability of the business). Therefore, the dynamic and temporal modeling of fairness criteria can improve the overall process. 
Creager et al. \cite{creager2020causal} propose a framework for causal modeling with the lending business as a foundation, demonstrating the utility of such an approach in simulating different scenarios in a profit-driven, but policy-constrained environment. The long-term vision of evaluating the implications of fairness on financial institutions and individuals seems to be a pursuit for researchers in an uncharted territory.

An auditor model is used to evaluate classification fairness, giving some constraints, and taking into account the explicability factor  \cite{hickey2020fairness}. However, even if the framework was tested on a credit risk dataset, the authors did not discuss the feasibility of the method in real-world terms, where the profitability of the lending business must be demonstrated along with fairness implementation.

The bias embedded in the data could be unobserved using an exclusively mathematical approach. Lee and Floridi \cite{lee2021algorithmic} argue for an approach in which the context -dependency of data is exploited. They compared different algorithms in terms of fairness capability, showing how some of the algorithms were not successful because of the association between the protected attributes and other features in the dataset (which are a proxy for the loan outcome). However, the study makes an unrealistic assumption by considering the lender’s intention to maximize the value of the loans. In practice, the overall profitability of the lending business should be considered (see \cite{petrides2022} for a profit-driven approach in credit scoring). In the same line of work, Kilbertus et al. \cite{kilbertus2020fair} discuss on the difficulties of achieving perfect fairness and maximizing profits in the context of potentially biased previous decisions. The credit scoring models are trained based on previous lending data, leading to sub-optimal performance. They proposed a stochastic decision rules system that attempts to improve decisions in terms of utility and fairness.

A group unfairness index was introduced by Szepannek and Lubke \cite{szepannek2021facing} to easily quantify and compare the fairness of the models. This index is based on group fairness by acceptance rate, which is a relevant fairness definition in the context of credit scoring.

\section{Methods}\label{sec4}

A pipeline with three different approaches can be depicted based on the literature on machine-learning fairness. The general consensus is to first measure the bias in algorithmic decisions by using one or several assessment metrics, and then to mitigate unfairness by employing a specific mitigation method. Finally, the results were compared in terms of the bias evaluation and performance metrics before and after the mitigation method was applied.  Figure~\ref{fig:pipeline} shows the fairness pipeline used in this study. Next, we present the bias assessment metrics used to evaluate the fairness of the classification throughout our experiments, and the bias mitigation algorithms tested in our benchmarking environment.

\subsection{Bias assessment metrics}
\label{sub:bias}
As shown in the literature review section, different criteria for evaluating the fairness of a classifier have recently been developed. Initially, they were built intuitively to address the unfairness of ML algorithms in classification problems. Subsequently, optimizations and new visions haveemerged(like using the economic inequality indexes \cite{speicher2018unified}), resulting in a set of measures generally accepted as a standard in the field.

\subsubsection{Separation metrics}

The \textit{average odds difference} \cite{hardt2016equality} is a measure of classification fairness   which computes the difference between the FPR for protected and unprotected groups and adds to it the difference between the TPR for protected and unprotected groups. This is especially useful in the credit scoring context, since the TPR (clients classified as goods and indeed repaying their debt) and FPR (clients wrongly classified as goods, not repaying their debt) influence the profitability of the business model.

\begin{equation}
\begin{split}
SP=& \frac{1}{2}\lvert(FPR_{(D=unprivileged)}-FPR_{(D=privileged})+\\
&+(TPR_{(D=unprivileged)}-TPR_{(D=privileged)})\rvert\label{eq4} 
\end{split}
\end{equation}

\begin{figure*}
\centering
\includegraphics[width=0.82\textwidth]{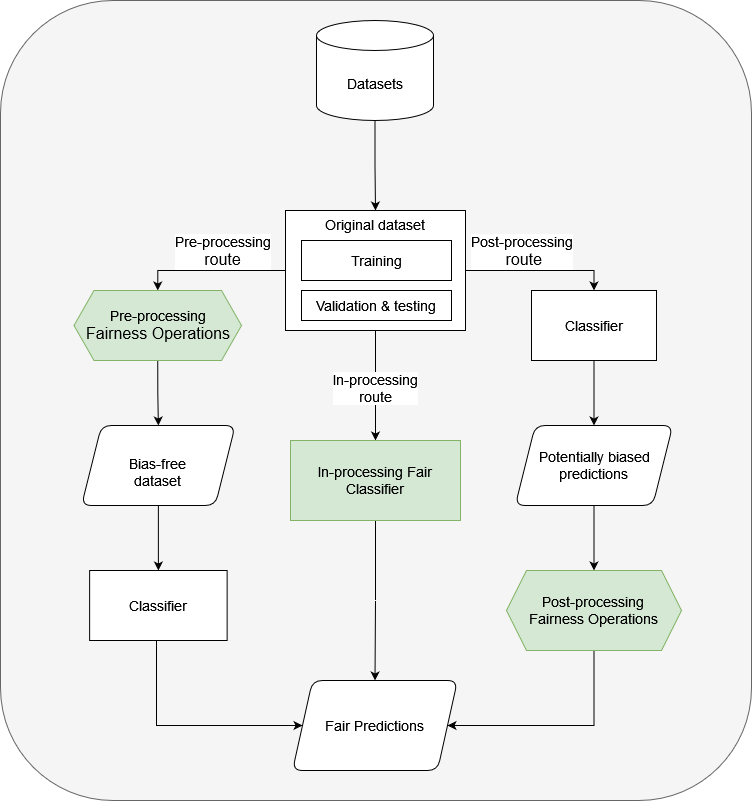}
\caption{\label{fig:pipeline}Fair processing pipeline}

\end{figure*}

A value close to 0 was obtained in the case of fair classification. For practical reasons we consider fairness to be inside the interval (-0.1,0.1).

\textit{Equal opportunity difference} is a relaxed version of the average odds difference, considering only the difference between the TPR for protected and unprotected groups. In the credit scoring context, this means that the classifier should have the same error rate when suggesting acceptance of loans in both protected and unprotected groups.
The requirement of equalized errors puts pressure on decision makers to improve the misclassification rates by optimizing models and increasing the quality of data \cite{barocas-hardt-narayanan}. The fairness interval considered for this metric is (-0.1, 0.1).

\begin{align}
EOD= TPR_{(D=unprivileged)}-TPR_{(D=privileged)}\label{eq5} 
\end{align}

\subsubsection{Independence metrics} 

\textit{Statistical parity difference} measures the difference between the probabilities of acceptance in the protected and unprotected groups. A value close to zeroimplies the same acceptance rate for both groups \cite{dworkfairness}. The fairness range for this metric is considered in the interval (-0.1,0.1).

\begin{equation}
\begin{split}
SPD=& Pr(\hat{Y}=1\rvert D=unprivileged)-\\
&-Pr(\hat{Y}=1\rvert D=privileged)\label{eq6}  
\end{split}
\end{equation}

\textit{Disparate impact} starts from the idea of independence and calculates the ratio between the probability of acceptance for unprivileged and  privileged groups. A value close to 1 implies an ideal degree of fairness, while values lower than 1 indicate an advantage for the privileged group and values higher than 1 indicate an advantage for the unprivileged group \cite{feldman}. In a more flexible approach, the interval (0.8, 1.25) was considered acceptable for a classifier to be considered fair.

\begin{align}
DI= \frac{Pr(\hat{Y}=1\rvert D=unprivileged)}{Pr(\hat{Y}=1\rvert D=privileged)}\label{eq7} 
\end{align}

\subsubsection{Individual fairness metric}

The \textit{Theil index} lies in the category of metrics measuring inequality in economics, which is a special case of the generalized entropy index ($\alpha$=1). This allows us,  in the context of fairness, to observe inequalities both at the group level (between groups) and at the individual level (within group) \cite{speicher2018unified}.

\begin{align}
Theil=\frac{1}{n} \sum_{i=1}^{n}\frac{b_i}{\mu}ln\frac{b_i}{\mu},
where \;  b_i=\hat{y_i}-y_i+1
\label{eq8} 
\end{align}

In formula \ref{eq8} \textit{n} represents the number of instances in the dataset; $\mu$ is the mean value of the benefits within a group; and $\hat{y}$ and $y$ are the individual predicted and true outcomes, respectively. A value close to 0 was used as proxy for fair learning. 

\subsection{Bias mitigation methods}
Bias mitigation methods are classified into three categories based on their position in a fair AI process (see Figure \ref{fig:pipeline}). The pre-processing methods address fairness issues before applying the classification method, and are therefore independent of the classification algorithm. To ensure fairness, in-processing methods have been developed within the classification method. The post-processing bias mitigation methods were placed at the end of the fair processing pipeline. They made adjustments after the model was trained, considering the protected attribute restrictions. Post-processing methods, such as pre-processing methods, are independent of the classification algorithm used, permitting a classifier agnostic approach to mitigate bias. The methods presented below were chosen from among those presented in highly cited studies on this topic.

\subsubsection{Pre-processing mitigation methods}

\textit{Reweighing} is an early stage technique in fair AI areas to mitigate bias. It is centered on the idea of reweighing data without relabeling to remove discrimination \cite{calders2009, kamiran2012data}. The algorithm attempts to achieve fairness by assigning lower weights to favored attributes.

The weights are assigned as the ratio between the expected and observed probabilities to see an instance with its protected attribute in a class:

\begin{align}
W(X)=\frac{P_{exp}(S=X(S)\wedge Class=X(Class))}{P_{obs}(S=X(S)\wedge Class=X(Class))}, 
\label{eq9} 
\end{align}
where X is the entire dataset, and S is a binary variable indicating whether an individual is a member of a protected group. Thus, the resulting dataset carries a fair representation of protected instances.

\textit{Learning fair representations} is a pre-processing technique that encodes data while obfuscating information about protected groups \cite{zemel2013learning}. The method acts as a clustering model, building prototypes based on the requirement that one element from a protected group is mapped to a certain prototype with the same probability as an element from an unprotected group (using the statistical parity criterion described in subsection bias assessment metrics). Formally, this is represented by

\begin{align}
P(Z=k\rvert x^+ \in X^+) = P(Z=k\rvert x^- \in X^-), \forall  k,
\label{eq10} 
\end{align}
where Z is a random attribute with k classes, each representing a prototype in the context of the clustering model; $x^+$ represents an individual from a protected group ($X^+$); and $x^-$ represents an individual from an unprotected group ($X^-$).

Another method developed by Feldman et al. \cite{feldman} is \textit{Disparate Impact Remover}. The algorithm detects the disparate impact (described in subsection Bias assessment metrics) and attempts to repair the data to achieve fairness. Repair is performed to preserve the predictability of the target variable and to preserve the relative per-attribute ordering (ranking). The synthetic dataset maintained the original values for the protected attribute(s) and target variables. A more relaxed version of the algorithm considers a trade-off between fairness and utility (accuracy) and performs partial repair. Depending on the problem addressed and the classifier performance, a compromise can be achieved.

\subsubsection{In-processing mitigation methods}
By adding \textit{adversarial learning} to the predictor, Zhang et al. \cite{zhang2018mitigating} introduced a framework in which the target is to increase the chances of the predictor for classification, while minimizing the adversary’s chances of predicting the protected feature. This method uses equality of odds as a fairness metric.

The term \textit{gerrymandering} in the context of AI fairness questions the ability of a fairness constraint applied to a certain protected group to ensure fairness at the individual level. This implies that an apparent fair classification at the group level does not necessarily imply fairness for all individuals. The \textit{Gerryfair} method \cite{pmlr-v80-kearns18a} consists of two algorithms that work as a learner and an auditor, working as a cost-sensitive classification oracle using linear methods. The fairness metrics used were false-positive rates, false-negative rates, and statistical parity.

Another cost-sensitive method for mitigating bias is the \textit{Exponentiated gradient reduction} \cite{agarwal2018reductions}. The algorithm consists of a sequence of two reductions aimed at yielding a classifier with the lowest error in the context of the defined constraints. The metrics considered for fairness optimization were statistical parity and equalized odds. This method was designed only for binary classifications.

Building on the work of Agarwal et al. \cite{agarwal2018reductions}, the \textit{Grid search reduction} was developed to predict continuous outcomes instead of discrete classification \cite{agarwal2019fair}. This direction was chosen because of the practical need for numerical prediction (e.g., quantifying the risk of default in the credit scoring setting). The fairness metrics used by the method are the statistical parity and bounded group loss, which is essentially the control of the prediction error for the protected group. A separate optimization algorithm was built for each metric.

One approach that attempts to satisfy several fairness metrics is the so-called \textit{Meta classifier} \cite{celis2019classification}. The central concept of this method is to develop an algorithm for a large family of classification problems. The current implementation supports only two metrics: the false discovery rate and  disparate impact. A very practical feature is the possibility of varying the constraint  ($\tau$) to achieve a reasonable trade-off between fairness and accuracy.

The \textit{prejudice remover} is one of the methods developed in the early stages of AI fairness evolution, proposing a regularization approach applied to logistic regression \cite{kamishima2012fairness}. It defines prejudice as the statistical dependence between protected attributes and other information. Further, it regularizes  learners' behavior regarding sensitive attributes by enforcing independence. The regularization parameter can be adjusted based on the accuracy-fairness trade-off.

\subsubsection{Post-processing mitigation methods}

One of the first methods developed for fairness post-processing was the \textit{reject ption classification} \cite{kamiran2012decision}. This method uses posterior probabilities from a classifier to label instances to neutralize discrimination. Rejected instances situated in a so-called critical region are assigned special labels depending on the protected group membership. These are considered to be easily influenced by bias. The method relables the instances using two cost-sensitive matrices for deprived and favored groups by optimizing loss functions  ($L$) for the two categories, according to a discrimination-accuracy trade-off coefficient $\theta$ (Eq.\ref{eq:roc}).
\begin{align}
L^d_{+,-} = L^f_{-,+} = \theta/(1-\theta)
\label{eq:roc} 
\end{align}

\textit{Calibrated equalized odds post-processing} is another method that changes the output labels after classification to preserve fairness. Introduced by Pleiss et al. \cite{pleiss2017fairness}, the method builds upon the work of Hardt et al. \cite{hardt2016equality}, who introduced the equalized odds fairness measure. Calibration is added to the method of mitigating bias by providing the possibility of ensuring fairness for both protected and unprotected groups without leaving the option of incentivizing the algorithm when considering the sensitive feature. The method gives the practitioner the freedom to choose the level of fairness constraint, an adjustment needed when the classification accuracy suffers after calibration.

\section{Experiments}
\label{sec:experiments}

To benchmark the different mitigation methods described above and evaluate them according to classification performance indicators and bias assessment metrics, we used two datasets. The first is the well-known German credit dataset available from the UCI Machine Learning Repository\footnote{ Source: https://archive.ics.uci.edu/ml/datasets/statlog+(german+credit+data)} \cite{Dua:2019}. The other dataset contains information on customers applying for personal loans, obtained from a Romanian bank\footnote{https://www.kaggle.com/datasets/zafish/consumer-loans}. 
We performed our experiments according to the pipeline described in Figure \ref{fig:pipeline} using the framework of AI Fairness 360 \cite{bellamy2019ai}.

\subsection{Data}

We used the German credit dataset, which is one of the most popular datasets used for benchmarking in the field and has also been included in other research on fair AI, such as the works of \cite{szepannek2021facing, lequy, kozodoi2022fairness}.
Real-world datasets in the field of credit scoring are rather rare and imply business-specific challenges (such as class imbalance and profit ratios). We therefore considered it useful to bring to the attention of the community a novel dataset containing consumer loan data. A summary of the statistics for both datasets is provided in Table \ref{tab:data}.

\begin{table*}[h]
\begin{center}
\begin{minipage}{320pt}
\caption{Summary statistics for the datasets}\label{tab:data}%
\begin{tabular}{@{}llllll@{}}
\hline
Dataset & Instances & Cat/Num/Bin & Default & Protected  & Protected \\&&Attributes&Ratio &Attribute & Group Ratio\\
\hline
German credit & 1000 & 32/7/1 & 30\% & Age &19\%\\
Consumer loans & 21568 & 7/7/9 &5.7\% & Age &4.09\%  \\
\hline
\end{tabular}
\end{minipage}
\end{center}
\end{table*}

Both datasets consist of samples of loan applications, with the target attribute being the outcome of the loan (good or bad). In the industry standard, clients repaying their loans are called \textit{goods}, while those not reimbursing the debt are called \textit{bads}.

The independent attributes in the datasets can be divided into three categories: \textit{sociodemographic} attributes, including information about age, education, profession, and \textit{customer history} providing information regarding the relationship between the customer and the bank (e.g., other products owned, previous loans); and \textit{economic information} such as client's income or loan amount.

A detailed view of the attributes of the consumer loan dataset is provided in Table \ref{tab:consumer} in Appendix \ref{secA1}.

We considered the \textit{age} of the applicant as a protected attribute, as suggested in other studies \cite{zhou2021improving, kozodoi2022fairness, lequy}. The threshold at the age of 25 years was considered to differentiate between the vulnerable group (under 25) and invulnerable group (25 and over) \cite{kamiran2009}. Moreover, when testing the importance of variables in relation to the target variable (Default Flag) for the consumer loan dataset, \textit{age} had the highest score, $-Log(p)=73.736$. The weight of evidence for this score suggests a split in the data at age 23. However, the threshold was maintained at $25$ for reasons of comparability with other  studies.   Another possible vulnerable attribute, \textit{marital status}, followed age in terms of importance, with a score of $56.937$.

Even if other works \cite{hurlin2021fairness} suggest the use of gender as a protected attribute, we consider it inappropriate, since the use of attributes such as gender and race is explicitly prohibited by laws worldwide, ensuring fairness through unawareness.

\subsection*{Experiment design}

We conducted our experiments starting with data pre-processing. As expected, the workload for cleaning the data was high in the case of the consumer loan dataset, whereas the German credit dataset was already curated.

\subsubsection*{Curating the consumer loans dataset}
We started by dropping irrelevant attributes such as ID, birthplace, and profession, the latter having too many different classes and making the attributes more noisy than useful.

For categorical attributes with missing values, we added a missing class. The behavior of attributes with missing values can be interesting to observe when the missing values have a certain significance. For example, a missing value for an attribute such as workplace seniority can be explained if the loan applicant is already retired. Other missing information can be related to data collection issues. The pre-processing of categorical attributes was finalized with one-hot encoding for the transformation into numerical values, a condition for being able to run all algorithms in the benchmarking phase.

Missing data for the numerical attributes was imputed with the median value for each attribute. This method was used because the distributions were skewed.

\subsubsection*{Experiments setup}
Next, for the consumer loans, the data was partitioned randomly into a training set (70\%), validation set (15\%), and test set (15\%), considering a stratification that assigns the instances to each set based on the target variable distribution. Because the German credit data is rather small (1000 instances) we partitioned the data into training (70\%) and testing (30\%).

After splitting the data, we verified how the split affected the difference in the mean outcomes (statistical parity) between the protected and unprotected groups. A large difference would indicate significantly different conditions between the training and testing environments. Table \ref{tab:parity} lists the values for the two datasets.

\begin{table}[h]
\begin{center}
\begin{minipage}{200pt}
\caption{Statistical parity differences for the training/validation/test datasets}\label{tab:parity}%
\begin{tabular}{@{}llll@{}}
\hline
Dataset & Training & Validation & Test\\
\hline
German credit & -0.1260 & N/A & -0.1335 \\
Consumer loans & -0.1971 & -0.2050 & -0.1997  \\
\hline
\end{tabular}

\end{minipage}
\end{center}
\end{table}

The protected group was defined considering the values of the attribute \textit{age}, with values less than 25. The favorable label for the target variable was set to represent loan applicants' good behavior. Note that the real-world consumer loans dataset encodes the target variable \textit{Default Flag}, meaning that the favorable label would be, in this case, 0. 

Next, we developed experiments in accordance with the pipeline shown in Figure \ref{fig:pipeline}. For each bias mitigation method, we chose the corresponding path according to the class it belonged to (pre-processing, in-processing, post-processing).

For the pre-processing and post-processing methods, we used logistic regression as the non-sensitive classifier to be trained. In the case of in-processing methods, we used the same classifiers to initially test the fairness as those used by the methods themselves. This is one of the disadvantages of in-processing debiasing methods. As mentioned in the Methodology section, this type of bias mitigation algorithm does not usually allow the user to select different classifiers for training because the methods are built around certain classifiers, which does not permit the same independence as the pre-processing and post-processing methods. As we will see in the results section, the outcomes may be quite different among the three categories.

To evaluate the performance of different bias mitigation metrics, we used fairness-specific metrics, a general classification metric, and a profit metric (for a synthetic view, see Table 3). The fairness-specific metrics are described in more detail in the methodology section. The general classification metric employed was \textit{balanced accuracy}, which represents the average of sensitivity and specificity  (see eq. \ref{eq11}). This method is particularly useful for imbalanced data sets. Even if a simple accuracy rate was used by most of the studies introducing fairness processing methods, we considered it insufficient in our context.
\begin{align}
\begin{split}
Balanced \ accuracy&= \frac{1}{2}(\frac{TP}{TP+FN}+\frac{TN}{TN+FP})\\
&=\frac{1}{2}(TPR+TNR)
\label{eq11} 
\end{split}
\end{align}

For profit calculations, we used the ROI measure \cite{VERBRAKEN2014505} to estimate the outcome of the correct classification of a good client. The value used in the experiments was ROI=0.34, and the loss was accounted  for at a rate  of 0.9. We determined these values by considering the specifics of the Romanian market during on-site discussions with the bank representatives. The ROI estimation follows the same process as reported by \cite{petrides2022} by observing the behavior of clients fully reimbursing their loans. Some of the loans are reimbursed earlier than initially scheduled, causing a lower than anticipated ROI. Therefore, the ROI was computed by multiplying the interest rate by the loan term and adjusting it by an \textit{early repayment coefficient (ERC)} (eq. \ref{eq12}).

\begin{align}
ROI= Interest \ rate\  \times \  \#years\  \times \ ERC
\label{eq12} 
\end{align}

The loss incurred by loans that are not repaid (false positives in the context of our experimental setup) does not necessarily mean a total loss, as the clients made some payments before default. Note that most application scorecards are built considering the probability of default during a one-year time horizon after failing to repay for 90 consecutive days. The \textit{loss coefficient (LC)} was set at 0.9.
The profit of the model was computed by adjusting the TPR with the ROI and subtracting the losses caused by FP misclassification.

\begin{align}
Profit= TPR  \times \  ROI  - \ FPR \times LC
\label{eq13} 
\end{align}

Because the German credit dataset does not have any information regarding the ROI, we used the same values for calculating profits for both datasets, a practice also observed in the work of Kozodoi et al. \cite{kozodoi2022fairness}
. Table \ref{tab:metrics} summarizes the metrics employed in evaluating fairness, classification accuracy and profit achieved by each method. The results provided represent the 10-fold cross-validation average for each measure.

\begin{table*}[h]
\begin{center}
\begin{minipage}{320pt}
\caption{Performance metrics}\label{tab:metrics}%
\begin{tabular}{@{}lll@{}}
\hline
Metric & Fairness interval & Description\\
\hline
Average odds difference & (-0.1, 0.1) & Separation metric \\
Equal opportunity difference & (-0.1, 0.1) & Separation metric \\
Statistical parity difference & (-0.1, 0.1) & Independence metric \\
Disparate impact & (0.8, 1.25) & Independence metric \\
Theil index & $\approx$ 0 & Individual fairness for independence \\
Balanced accuracy & N/A & Classification performance metric \\
Profit & N/A & Model's profit (\%) \\
\hline
\end{tabular}
\end{minipage}
\end{center}
\end{table*}

\section{Results and Discussion}

This section presents the results obtained after running our experiments on the two datasets, the best values for each criterion are underlined (see Tables \ref{tab:results1} and \ref{tab:results2}). 

By analyzing the results, we can denote the overall loss of accuracy and profit when bias is mitigated for all fairness processors.

In several cases, our findings contradict our expectations (based on the literature review) that all fairness constraints cannot be satisfied at the same time. In the case of the consumer loan dataset, methods such as \textit{Learning Fair Representations, Disparate impact remover}, and\textit{Exponentiated Gradient reduction} managed to achieve fairness for all five metrics. For German credit data, \textit{Grid Search Reduction} achieved comparable performance. 

Two methods failed to achieve consistent results for each tested dataset. As will be shown later, this may have been caused by bias in the methods.

The mixed results obtained after applying the methods to the two datasets show a relevant connection between the method and data quality. For example, only one method managed to reduce the Theil index significantly in the case of the German credit dataset, which could be a consequence of the small amount of data (1000 instances). The consumer loan dataset was very imbalanced (5.7\% defaulted loans), causing accuracy classification problems. In this context, the use of \textit{Balanced Accuracy} is particularly important in the context of imbalanced data for differentiating classification performance. 

Figures \ref{fig:2} and \ref{fig:3} in Appendix \ref{secB1} provide a visual representation of the biased and de-biased values across the multiple mitigation methods. Each sub-chart displays the relationship between the balanced accuracy and a specific fairness measure, highlighting the impact of different mitigation methods on model fairness and accuracy. The effectiveness of the  mitigation methods in reducing bias can be visualized, by observing the grey areas representing the desired range for fairness, highlighting the ideal performance zone.

Next, we review the performance of each fairness processor and discuss the difficulties and special processing required during the experiments.

The pre-processing method of \textit{Reweighing} the examples in each group clearly performs well for most of the bias indicators. The classification performance is not significantly affected, which is an advantage over other more sophisticated methods, but a value higher than the recommended value of the Theil index might suggest unfairness at the individual level. In the case of the German dataset, although the algorithm is improving without any doubt the fairness metrics, the results are very volatile owing to the small amount of data sample.   

While the \textit{Learning Fair Representations} method might seem difficult to set up because of the various values and combinations of the parameters (two fairness parameters, classification threshold, and loss tolerance), the results were promising for both datasets. Because this is a pre-processing bias mitigation method, the operator has the freedom to choose more powerful classifiers.

When applying the \textit{Disparate Impact Remover}, the algorithm optimizes the value of the disparate impact but with a high cost for accuracy and profits, leaving no room for compromise between fairness and accuracy or profits.

\begin{table*}[h]
\begin{center}
\begin{minipage}{345pt}
\caption{Benchmark results (Consumer loan dataset)}\label{tab:results1}%
\begin{tabular}{@{}llllllllll@{}}
\hline
\textbf{Fairness}& \textbf{Proc.} & DI & SP & AOD & EOD & TI &  BAcc & P \\
 \textbf{processor}& \textbf{type} &  &  &  &  &  &  &\\
\hline
Reweighing & Pre & 0.818 & -0.127 & -0.026 & -0.148 & 0.303 &  0.764 & 0.319 \\
 &  &  &  &  &  &  &  & \\ 
 Learning Fair  & Pre & 0.850 & -0.044& -0.009 & -0.050& 0.087 & 0.578 & 0.283 \\
 Representations& &  &  &  &  &  &    &\\ 
 Disparate  & Pre & 0.964 & -0.035 & -0.059 & -0.011 & 0.019 & 0.507 & 0.270 \\
impact remover & &  &  &  &  &  &  &  &\\ 
 Optimized  & Pre & 0.000 & -0.335 & -0.284 & -0.344 & 0.962  & 0.566 & 0.278\\
pre-processing&  & &  &  &  &  &  &  &\\ 
 Adversarial&  In &  0.868 & -0.128 & -0.054 & -0.086 & 0.025 &  0.645&0.289\\
 Debiasing&  & &  &  &  &  &  &  & \\ 
 GerryFair&  In &  0.830 & -0.169 & -0.205 &-0.131 & 0.022  & 0.522&0.272\\
 & &  &  &  &  &  &  &  &\\ 
Meta Classifier*&  In &  0.762 & -0.048 & -0.004 & -0.010 & 0.075  & 0.619 &0.288\\
 & &  &  &  &  &  &  &  &\\ 
 Exponentiated&  In &  0.918 & -0.077 & -0.003 &-0.007 & 0.058 &  0.559 &0.277\\
 Gradient Red. &  & &  &  &  &  &  &  & \\ 
  Grid Search&  In &  1.036 & 0.029 & 0.117 & 0.04 & 0.205 &  0.560 &0.279\\
 Reduction &  & &  &  &  &  &  &  & \\ 
Prejudice  &  In &  N/A & N/A & N/A & N/A & 0.058 & 0.500 & 0.269\\
Remover* & &  &  &  &  &  &  &  &\\ 
 Reject Option &  Post &  0.984 & -0.010 & 0.153 & 0.030 & 0.313 &  0.714 &0.313\\
 Classification &  & &  &  &  &  &  &  & \\ 
 Calibrated  &  Post &  1.036 & 0.029 & 0.117 & 0.04 & 0.205 &  0.560 &0.279\\
 Odds-Equalizing &  & &  &  &  &  &  &  & \\

No bias &  N/A &  0.184 & -0.591 & -0.395 &-0.574 & 0.290  & 0.776& 0.321\\ 
mitigation &  & &  &  &  &  &  &  & \\
\hline
\end{tabular}
\footnotetext{ Abbreviations: DI = disparate impact; SP = statistical parity; \\AOD = average odds difference; EOD = equal opportunity difference;\\ TI = Theil index; BAcc = balanced accuracy; P = profit.\\ * unstable results}

\end{minipage}
\end{center}
\end{table*}

The \textit{Optimized Pre-processing} algorithm was found to be extremely slow when dealing with a large number of attributes. We had to reduce the number of attributes by selecting only the top 5 attributes considering information value, along the protected attribute and the target attribute, to create a testing environment similar to that described by Calmon et al. \cite{calmon2017optimized}. Some of the fairness metrics remained at unwanted levels, whereas accuracy decreased significantly

\textit{Adversarial debiasing} is implemented using tensor flow for classification, employing a predictor and adversary model to achieve fairness. The method achieves fairness in almost all the chapters, with a loss in accuracy. The slightly better results compared to the other methods might be due to the use of tensor flow instead of logistic regression.

The \textit{GerryFair} algorithm is an in-processing method for mitigating bias in datasets that considers the individual level of unfairness. The dataset needs to be specially pre-processed in order to fit its requirements in a manner similar to the \textit{optimized pre-processing} method. It also has several hyper-parameters to be tuned before being able to perform a suitable job for the dataset.These parameters include the fairness target ($\gamma$), maximum number of iterations, the maximum L1-Norm to be used for dual variables and the learner to be used. The learner must be a regressor. Our typical logistic regression used as a classifier in the experiments was not supported by this method; therefore we tried the learners recommended by Kearns et al. \cite{pmlr-v80-kearns18a}. We found it difficult to tune the parameters in order to obtain a reasonable trade-off between fairness and balanced accuracy. This method tends to overfit when dealing with the German dataset because of its small size.

\begin{table*}[h]
\begin{center}
\begin{minipage}{345pt}
\caption{Benchmark results (German credit dataset)}\label{tab:results2}%
\begin{tabular}{@{}llllllllll@{}}
\hline
\textbf{Fairness}& \textbf{Proc.} & DI & SP & AOD & EOD & TI &  BAcc & P \\
 \textbf{processor}& \textbf{type} &  &  &  &  &  &  &\\
\hline
Reweighing & Pre & 0.756 & -0.14 & -0.095 & -0.053 & 0.288 & 0.704 & 0.160 \\
 &  &  &  &  &  &  &  & \\ 
 Learning Fair  & Pre & 0.801 & -0.048  & -0.075  & -0.127 & 0.277 & 0.553 &  0.004  \\
 Representations& &  &  &  &  &  &  &  &\\ 
Disparate  & Pre &  0.819 & -0.148& -0.103 & -0.103 & 0.114 & 0.674 & 0.081\\
impact remover & &  &  &  &  &  &  &  &\\ 
Optimized  & Pre & 0.542 & 0.168 & 0.1044 & 0.002 & 0.221 & 0.642 & 0.067\\
pre-processing&  & &  &  &  &  &  &  &\\ 
 Adversarial&  In & 0.994 & -0.003 & 0.071 &-0.039 & 0.139 & 0.681 & 0.076  \\
 Debiasing&  & &  &  &  &  &  &  & \\ 
 GerryFair&  In &  0.811 & -0.156 & -0.121 & -0.097 & 0.12 & 0.655 & 0.068\\
 & &  &  &  &  &  &  &  &\\ 
Meta Classifier*&  In & 0.648 & 0.145 & 0.083 & 0.029 & 0.182 & 0.689 & 0.107  \\
 & &  &  &  &  &  &  &  &\\ 
 Exponentiated&  In & 0.8335 & -0.139 & -0.101 & -0.071 & 0.108 & 0.668 & 0.075 \\
 Gradient Red. &  & &  &  &  &  &  &  & \\ 
  Grid Search&  In & 0.937 & -0.052 & -0.027 & 0.076 & 0.094 & 0.677 & 0.080 \\
 Reduction &  & &  &  &  &  &  &  & \\ 
Prejudice  &  In & 0.719 & -0.237 & -0.196 & -0.193 & 0.112 & 0.454 & 0.074  \\
 Remover* & &  &  &  &  &  &  &  &\\ 
 Reject Option &  Post &  0.944 & -0.040 & 0.022 & -0.006 & 0.145 & 0.711 & 0.119\\
 Classification &  & &  &  &  &  &  &  & \\ 
 Calibrated  &  Post & 0.478 & -0.474& -0.483 & -0.321 & 0.115 &0.621 & 0.043 \\
 Odds-Equalizing &  & &  &  &  &  &  &  & \\

No bias &  N/A & 0.590 & -0.256 & -0.195 & -0.224 & 0.255 & 0.712 & 0.157  \\ 
mitigation &  & &  &  &  &  &  &  & \\
\hline
\end{tabular}
\footnotetext{ Abbreviations: DI = disparate impact; SP = statistical parity; \\AOD = average odds difference; EOD = equal opportunity difference;\\ TI = Theil index; BAcc = balanced accuracy; P = profit.\\ * unstable results}

\end{minipage}
\end{center}
\end{table*}

When using \textit{Meta Fair Classifier}, the user defines the importance of the fairness metric (false discovery ratio or disparate impact) as one of the inputs for the algorithm. The classifier showed promising results, but was not sufficiently stable. When running the algorithm multiple times on both datasets, the stability issue makes the option of averaging the results unusable, as sometimes fairness constraints are not achieved. The stabilization problem was also discussed by \cite{huang2019stable} and \cite{friedler2019comparative} in their work.

When testing the \textit{Exponentiated Gradient Reduction} algorithm, several fairness constraints were considered as parameters (Equalized odds, True Positive Rate Difference, Demographic Parity and Error Rate Ratio). Among these, the True Positive Rate Difference was associated with the best balanced accuracy and profit, while the fairness metrics achieved the targeted values for each constraint used in the case of the consumer loan dataset. . However, for the German credit dataset, the best results considering the constraints were obtained when Demographic Parity was used as a parameter. This shows how tied the results are to the data quality, and stresses the importance of experimenting when adopting a solution.

\textit{Grid Search Reduction} allows the user (along with the fairness constraint) to define the constraint weight to achieve a reasonable compromise between accuracy and fairness. This proves to be a convenient feature for an in-processing algorithm, allowing the practitioner to experiment and decide on the most convenient solution. However, the transition from unfair to fair classification does not seem to be smooth enough to have many options for a compromise between accuracy and fairness. The results showed a significant penalty in accuracy when fairness was achieved.

The last tested in-processing method, \textit{Prejudice remover} failed to achieve fairness on both datasets or became overfit. None of the variations in the fairness parameter of the algorithm ameliorated the results. A possible cause was reported by Kamishima et al. \cite{kamishima2013independence} to be in the design of the method, which is based on a hypothetical distribution. 

The post-processing method \textit{Reject Option Classification} showed some of the best results in terms of balanced accuracy and profit in a fairness constrained context. The fairness constraints allowed include statistical parity difference, average odds difference, and equal opportunity difference. The relatively high values of the Theil index indicate that unfairness persists at the individual level.

\textit{The Calibrated Odds-Equalizing post-processing} algorithm has been struggling to mitigate bias in the consumer loan dataset. It was tested with full and reduced data (feature selection applied); however, the fairness metrics did not show significant improvements over the unrestricted setup. However, the balanced accuracy was reduced after post-processing, in the case of consumer loans, from $ 0.755 $ to approximately $ 0.56 $, depending on the parameters applied to the algorithm (e.g., cost constraints can be chosen from the FNR, TNR, or weighted).

\section{Conclusion}\label{sec13}

This study adds to the literature on fair AI decision-making by benchmarking 12 bias mitigation methods against five fairness metrics and evaluating them in the context of credit scoring data, both in terms of balanced accuracy and profits. We based our experiments on two datasets, the classical German credit dataset and a novel consumer loans dataset from a Romanian bank, to show the challenges in implementing these methods in a real-world setup.

Almost every bias mitigation method benchmarked in our study managed to increase overall fairness in a potentially automated decision context. Considering the 5 fairness metrics covering (virtually) the entire spectrum of fairness definitions, we identified the strengths and weaknesses of each method. None of the fairness processors can be considered a leader in this area or a universal panacea for treating unfairness, while simultaneously satisfying the accuracy and profit criteria. For this reason, for a practitioner willing to mitigate bias in their decision, a group of methods should be employed and the most convenient one should be chosen at a later stage, based on cost limitations.
 
However, contrary to our expectations raised by the literature review that a fairness processor cannot satisfy all the fairness criteria, we have found several methods that were able to reduce unfairness in each chapter but incurred significant costs.

When tested on a real-world consumer loan dataset, some of the bias mitigation methods underformed in accuracy compared to the results reported by the studies introducing them. We also noted that the vast majority of these studies considered simple accuracy as the performance criterion, which in our case was unusable because of the highly class imbalanced data. This shows the difficulty of implementing fairness constraints in a real-world environment with highly imbalanced data (the typical scenario for credit risk analysis), where the loss in accuracy can translate to severely increased costs or diminished profits.

Our methodology describes the differences in the ways input data needs to be prepared before feeding the different methods: different encoding of the sensitive attributes, the necessity to transform categorical features by hot-encoding them, or the importance of reducing the number of features in the dataset because of convergence time.
One of the limitations of our work is that it is observational and limited to case studies, not being able to provide answers regarding the causes that may generate fairness gaps. This could be the object of a qualitative (rather than quantitative) study of the lending business, which looks outside the statistical definitions of fairness.

The accuracy results provided in this study can be optimized by employing different classifiers in a case-by-case manner. However, this optimization was not included in our scope, with the focus being on the bias processors and fairness metrics. We used logistic regression to ensure that the methods were truly comparable, as most in-processing models use this algorithm for classification, which is also widely known as the industry standard  \cite{XIAO2021508, GUNNARSSON2021292}. Wherever possible, after choosing one or several bias mitigation methods, a practitioner should test several classifiers to achieve the best results.

In the wake of cost-sensitive classification methods \cite{verbeke2022or} aiming to improve the effectiveness of different processes, among which is the lending business, the addition of fairness constraints to a cost-sensitive framework for credit scoring could be a possible solution to the still open challenge of managing the trade-off between profit, risk and fairness. 

Because some bias mitigation methods are computationally  intensive, one may consider benchmarking them by calculating the associated energy and carbon footprint, which is becoming of interest in the field of data science \cite{patterson2022}. In conjunction with the other results, this may be used as a criterion for deciding the bias processor to be used in practice.

\appendix[Appendix]
\section{A}
\label{secA1}
\hfill \\
Table \ref{tab:consumer} lists and describes the features of the consumer loans dataset.
\hfill \\

\begin{table*}[h]
\caption{\label{tab:consumer} Description of features for the consumer loans dataset}
\centering
\begin{tabular}{c  l  c  c c}
\hline
\# & Name & Type & Domain & Description \\
\hline

1 & Product& Categorical & \#6 & Different consumer loan types\\
2 & Age & Numerical & $\mathbf{R}^+$  & Age of the applicant \\
3 & Area & Categorical & \#4 & Urban/ Rural area \\
4 & Residential Status & Categorical & \#5 & Residence type (e.g. property, rented)\\ 
5 & Education & Categorical & \#10 & Level of education \\
6 & Marital status & Categorical & \#4 & Personal status \\
7 & Household members & Numerical & $\mathbf{R}^+$ & Number of family members in the household \\
8 & No. of dependents & Numerical & $\mathbf{R}^+$ & Number of dependents in the household  \\
9 & Income & Numerical & $\mathbf{R}^+$ & Monthly income of the applicant  \\
10 & Work seniority & Numerical & $\mathbf{R}^+$ & Work seniority in \# of years\\
11 & Business age & Numerical & $\mathbf{R}^+$ & Years active for the employer\\
12 & Legal form & Categorical & \#15 & Different types of business of the employer\\
13 & Economic sector & Categorical & \#18 & Activity sector of the employer\\
14 & Empolyees no & Categorical & \#9 & Company size for the employer\\
15 & Length relationship & Numerical & $\mathbf{R}^+$ &Relationship length between the applicant and the financial institution\\
16 &Debit card & Binary & \#2 & Other products with the financial institution \\
17 &Current account & Binary & \#2 & Other products with the financial institution \\
18 &Savings account & Binary & \#2 & Other products with the financial institution \\
19 & Salary account & Binary & \#2 & Other products with the financial institution \\
20 & Foreign account & Binary & \#2 & Other products with the financial institution \\
21 & Deposit & Binary & \#2 & Indicates whether the applicant has a deposit with the fin. institution \\
22 & Pension funds & Binary & \#2 & Pension funds subscriptions  \\
23 & Finalized loan & Binary & \#2 & Indicates whether the applicant has previously completed a loan. \\
24 & Default flag & Binary & \#2 & 1 - Defaulted loan; 0 - Non-defaulted loan  \\

\hline
\end{tabular}
\end{table*}

\section{B}
\label{secB1}

\begin{figure*}

  \centering
  \subfigure{\includegraphics[width=0.45\linewidth]{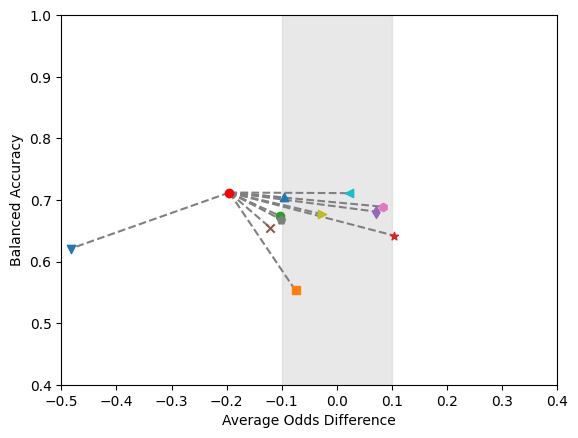}}
  \subfigure{\includegraphics[width=0.45\linewidth]{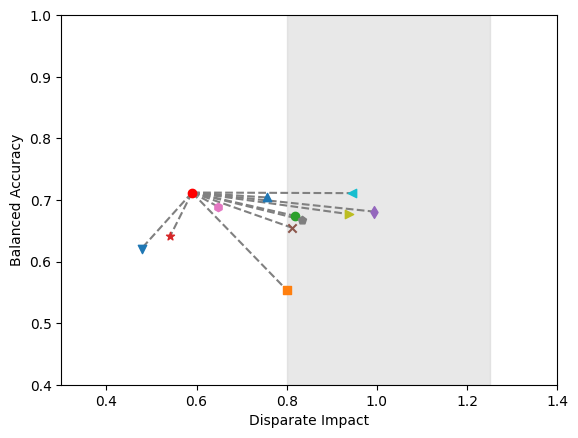}}

  \subfigure{\includegraphics[width=0.45\linewidth]{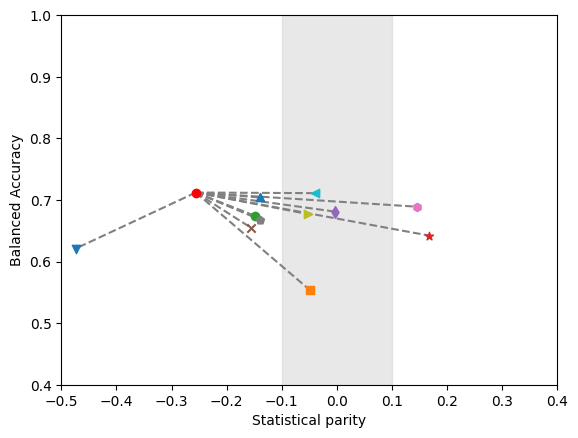}}
  \subfigure{\includegraphics[width=0.45\linewidth]{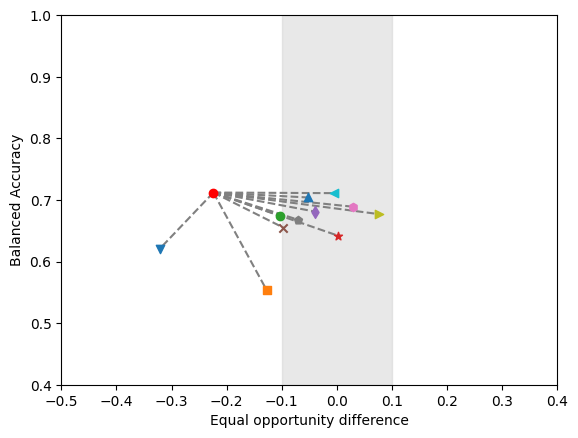}}
  
  \subfigure{\includegraphics[width=0.45\linewidth]{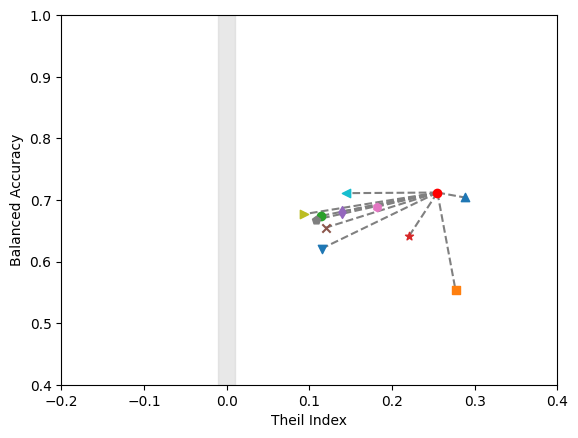}}
  \subfigure{\includegraphics[width=0.45\linewidth]{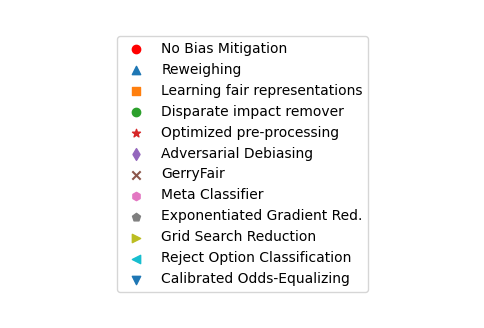}}
  
  \caption{Performances of bias mitigation methods considering the 5 fairness metrics on the german credit dataset (fairness intervals are in gray)}
  \label{fig:2}
\end{figure*}

\begin{figure*}
  \centering
  \subfigure{\includegraphics[width=0.45\linewidth]{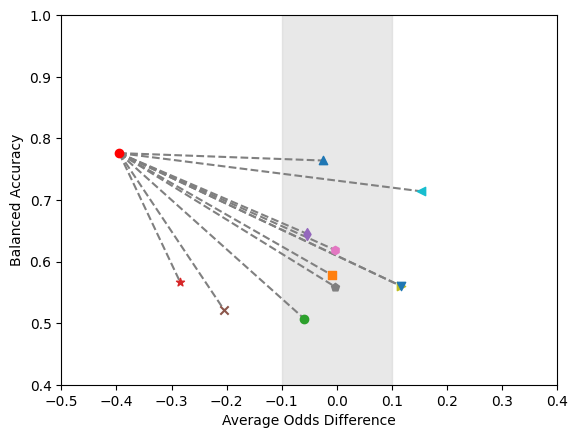}}
  \subfigure{\includegraphics[width=0.45\linewidth]{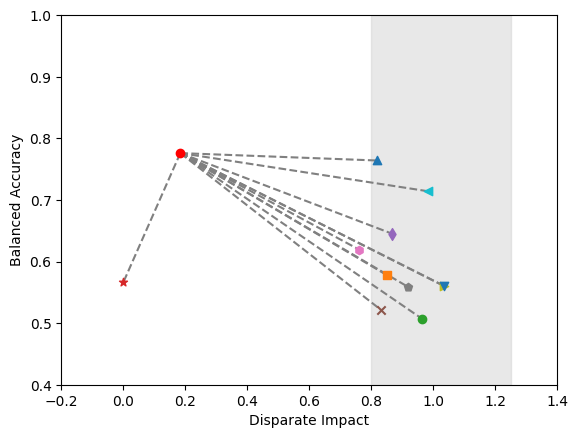}}

  \subfigure{\includegraphics[width=0.45\linewidth]{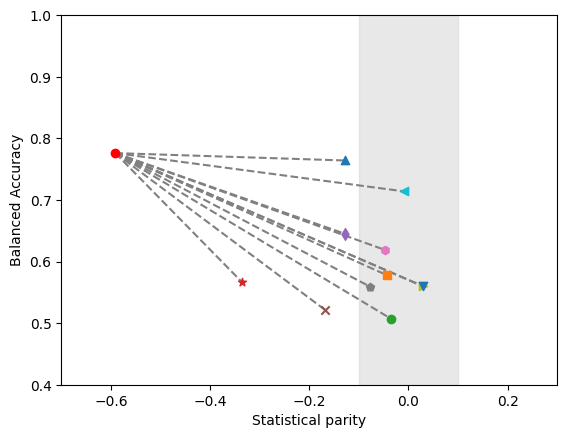}}
  \subfigure{\includegraphics[width=0.45\linewidth]{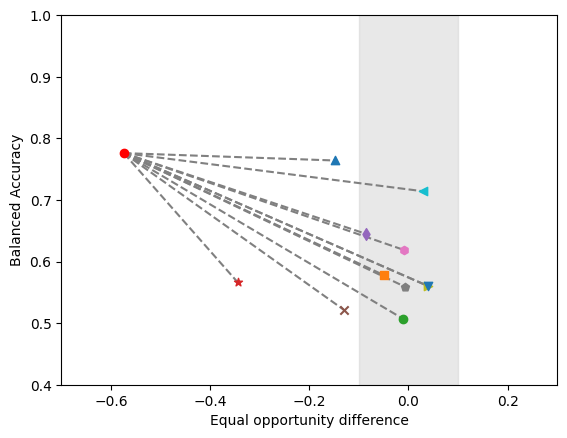}}
  
  \subfigure{\includegraphics[width=0.45\linewidth]{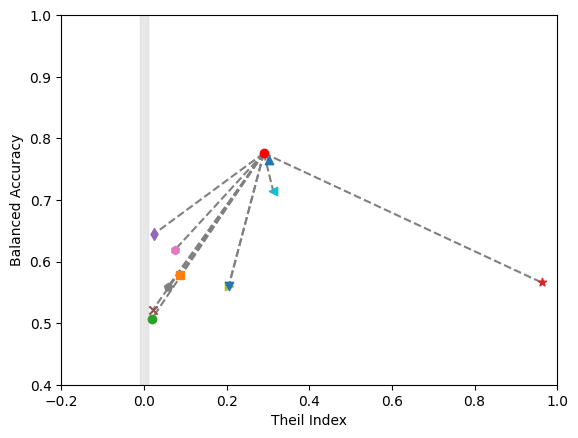}}
  \subfigure{\includegraphics[width=0.45\linewidth]{legend.png}}
  
  \caption{Performances of bias mitigation methods considering the 5 fairness metrics on the consumer loan dataset (fairness intervals are in gray)}
  \label{fig:3}
\end{figure*}

The charts in Figure \ref{fig:2} and \ref{fig:3} provide a visual representation of the debiased and not debiased values across multiple mitigation methods for both datasets. 

\bibliographystyle{unsrt}
\bibliography{bibliography}

\begin{thebibliography}{10}

\bibitem{agarwal2021ai}
Akshat Agarwal, Charu Singhal, and Renny Thomas.
\newblock Ai-powered decision making for the bank of the future.
\newblock {\em McKinsey \& Company.--2021.--March.--URL: https://www. mckinsey.
  com/\~{}/media/mckinsey/industries/financial\% 20services/our\%
  20insights/ai\% 20powered\% 20decision\% 20making\% 20for\% 20the\% 20bank\%
  20of\% 20the\% 20future/ai-powered-decision-making-forthe-bank-of-the-future.
  pdf (15.04. 2021)}, 2021.

\bibitem{blattner2021unpacking}
Laura Blattner, Scott Nelson, and Jann Spiess.
\newblock Unpacking the black box: Regulating algorithmic decisions.
\newblock {\em arXiv preprint arXiv:2110.03443}, 2021.

\bibitem{hurlin2021fairness}
Christophe Hurlin, Christophe P{\'e}rignon, and S{\'e}bastien Saurin.
\newblock The fairness of credit scoring models.
\newblock {\em Available at SSRN 3785882}, 2021.

\bibitem{neverending}
Federico Ferretti.
\newblock The {N}ever-{E}nding {E}uropean {C}redit {D}ata {M}ess.
\newblock Technical Report BEUC-X-2017-111, The European Consumer Organisation,
  Brussels, Belgium, October 2017.

\bibitem{european2020white}
European Commission.
\newblock White paper on artificial intelligence: A european approach to
  excellence and trust.
\newblock {\em Com (2020) 65 Final}, 2020.

\bibitem{un2021}
Human~Rights Council.
\newblock The right to privacy in the digital age.
\newblock {\em U.N. Doc. A/HRC/48/31}, 2021.

\bibitem{barocas-hardt-narayanan}
Solon Barocas, Moritz Hardt, and Arvind Narayanan.
\newblock {\em Fairness and Machine Learning}.
\newblock fairmlbook.org, 2019.

\bibitem{dworkfairness}
Cynthia Dwork, Moritz Hardt, Toniann Pitassi, Omer Reingold, and Richard Zemel.
\newblock Fairness through awareness.
\newblock In {\em Proceedings of the 3rd Innovations in Theoretical Computer
  Science Conference}, ITCS '12, page 214–226, New York, NY, USA, 2012.
  Association for Computing Machinery.

\bibitem{hardt2016equality}
Moritz Hardt, Eric Price, and Nati Srebro.
\newblock Equality of opportunity in supervised learning.
\newblock {\em Advances in neural information processing systems}, 29, 2016.

\bibitem{dieterich2016compas}
William Dieterich, Christina Mendoza, and Tim Brennan.
\newblock Compas risk scales: Demonstrating accuracy equity and predictive
  parity.
\newblock {\em Northpointe Inc}, 7(4), 2016.

\bibitem{simoiu2017problem}
Camelia Simoiu, Sam Corbett-Davies, and Sharad Goel.
\newblock The problem of infra-marginality in outcome tests for discrimination.
\newblock {\em The Annals of Applied Statistics}, 11(3):1193--1216, 2017.

\bibitem{kleinberg2017inherent}
Jon Kleinberg, Sendhil Mullainathan, and Manish Raghavan.
\newblock Inherent trade-offs in the fair determination of risk scores.
\newblock In {\em 8th Innovations in Theoretical Computer Science Conference
  (ITCS 2017)}, volume~67, page~43. Schloss Dagstuhl--Leibniz-Zentrum fuer
  Informatik, 2017.

\bibitem{paassen2019dynamic}
Benjamin Paa{\ss}en, Astrid Bunge, Carolin Hainke, Leon Sindelar, and Matthias
  Vogelsang.
\newblock Dynamic fairness-breaking vicious cycles in automatic decision
  making.
\newblock In {\em Proceedings of the 27th European Symposium on Artificial
  Neural Networks (ESANN 2019)}, 2019.

\bibitem{bono2021algorithmic}
Teresa Bono, Karen Croxson, and Adam Giles.
\newblock Algorithmic fairness in credit scoring.
\newblock {\em Oxford Review of Economic Policy}, 37(3):585--617, 2021.

\bibitem{parkes2019algorithmic}
David~C Parkes, Rakesh~V Vohra, et~al.
\newblock Algorithmic and economic perspectives on fairness.
\newblock {\em arXiv preprint arXiv:1909.05282}, 2019.

\bibitem{castelnovo2022clarification}
Alessandro Castelnovo, Riccardo Crupi, Greta Greco, Daniele Regoli,
  Ilaria~Giuseppina Penco, and Andrea~Claudio Cosentini.
\newblock A clarification of the nuances in the fairness metrics landscape.
\newblock {\em Scientific Reports}, 12(1):1--21, 2022.

\bibitem{kozodoi2022fairness}
Nikita Kozodoi, Johannes Jacob, and Stefan Lessmann.
\newblock Fairness in credit scoring: Assessment, implementation and profit
  implications.
\newblock {\em European Journal of Operational Research}, 297(3):1083--1094,
  2022.

\bibitem{feldman}
Michael Feldman, Sorelle~A. Friedler, John Moeller, Carlos Scheidegger, and
  Suresh Venkatasubramanian.
\newblock Certifying and removing disparate impact.
\newblock In {\em Proceedings of the 21th ACM SIGKDD International Conference
  on Knowledge Discovery and Data Mining}, KDD '15, page 259–268, New York,
  NY, USA, 2015. Association for Computing Machinery.

\bibitem{pmlr-v80-kearns18a}
Michael Kearns, Seth Neel, Aaron Roth, and Zhiwei~Steven Wu.
\newblock Preventing fairness gerrymandering: Auditing and learning for
  subgroup fairness.
\newblock In Jennifer Dy and Andreas Krause, editors, {\em Proceedings of the
  35th International Conference on Machine Learning}, volume~80 of {\em
  Proceedings of Machine Learning Research}, pages 2564--2572. PMLR, 10--15 Jul
  2018.

\bibitem{speicher2018unified}
Till Speicher, Hoda Heidari, Nina Grgic-Hlaca, Krishna~P Gummadi, Adish Singla,
  Adrian Weller, and Muhammad~Bilal Zafar.
\newblock A unified approach to quantifying algorithmic unfairness: Measuring
  individual \&group unfairness via inequality indices.
\newblock In {\em Proceedings of the 24th ACM SIGKDD international conference
  on knowledge discovery \& data mining}, pages 2239--2248, 2018.

\bibitem{fuster2022predictably}
Andreas Fuster, Paul Goldsmith-Pinkham, Tarun Ramadorai, and Ansgar Walther.
\newblock Predictably unequal? the effects of machine learning on credit
  markets.
\newblock {\em The Journal of Finance}, 77(1):5--47, 2022.

\bibitem{zarsky2016trouble}
Tal Zarsky.
\newblock The trouble with algorithmic decisions: An analytic road map to
  examine efficiency and fairness in automated and opaque decision making.
\newblock {\em Science, Technology, \& Human Values}, 41(1):118--132, 2016.

\bibitem{hurley2016credit}
Mikella Hurley and Julius Adebayo.
\newblock Credit scoring in the era of big data.
\newblock {\em Yale JL \& Tech.}, 18:148, 2016.

\bibitem{liu2018delayed}
Lydia~T Liu, Sarah Dean, Esther Rolf, Max Simchowitz, and Moritz Hardt.
\newblock Delayed impact of fair machine learning.
\newblock In {\em International Conference on Machine Learning}, pages
  3150--3158. PMLR, 2018.

\bibitem{creager2020causal}
Elliot Creager, David Madras, Toniann Pitassi, and Richard Zemel.
\newblock Causal modeling for fairness in dynamical systems.
\newblock In {\em International Conference on Machine Learning}, pages
  2185--2195. PMLR, 2020.

\bibitem{hickey2020fairness}
James~M Hickey, Pietro G~Di Stefano, and Vlasios Vasileiou.
\newblock Fairness by explicability and adversarial shap learning.
\newblock In {\em Joint European Conference on Machine Learning and Knowledge
  Discovery in Databases}, pages 174--190. Springer, 2020.

\bibitem{lee2021algorithmic}
Michelle Seng~Ah Lee and Luciano Floridi.
\newblock Algorithmic fairness in mortgage lending: from absolute conditions to
  relational trade-offs.
\newblock {\em Minds and Machines}, 31(1):165--191, 2021.

\bibitem{petrides2022}
George Petrides, Darie Moldovan, Lize Coenen, Tias Guns, and Wouter Verbeke.
\newblock Cost-sensitive learning for profit-driven credit scoring.
\newblock {\em Journal of the Operational Research Society}, 73(2):338--350,
  2022.

\bibitem{kilbertus2020fair}
Niki Kilbertus, Manuel~Gomez Rodriguez, Bernhard Sch{\"o}lkopf, Krikamol
  Muandet, and Isabel Valera.
\newblock Fair decisions despite imperfect predictions.
\newblock In {\em International Conference on Artificial Intelligence and
  Statistics}, pages 277--287. PMLR, 2020.

\bibitem{szepannek2021facing}
Gero Szepannek and Karsten L{\"u}bke.
\newblock Facing the challenges of developing fair risk scoring models.
\newblock {\em Frontiers in artificial intelligence}, 4, 2021.

\bibitem{calders2009}
Toon Calders, Faisal Kamiran, and Mykola Pechenizkiy.
\newblock Building classifiers with independency constraints.
\newblock In {\em 2009 IEEE International Conference on Data Mining Workshops},
  pages 13--18, 2009.

\bibitem{kamiran2012data}
Faisal Kamiran and Toon Calders.
\newblock Data preprocessing techniques for classification without
  discrimination.
\newblock {\em Knowledge and information systems}, 33(1):1--33, 2012.

\bibitem{zemel2013learning}
Rich Zemel, Yu~Wu, Kevin Swersky, Toni Pitassi, and Cynthia Dwork.
\newblock Learning fair representations.
\newblock In {\em International conference on machine learning}, pages
  325--333. PMLR, 2013.

\bibitem{zhang2018mitigating}
Brian~Hu Zhang, Blake Lemoine, and Margaret Mitchell.
\newblock Mitigating unwanted biases with adversarial learning.
\newblock In {\em Proceedings of the 2018 AAAI/ACM Conference on AI, Ethics,
  and Society}, pages 335--340, 2018.

\bibitem{agarwal2018reductions}
Alekh Agarwal, Alina Beygelzimer, Miroslav Dud{\'\i}k, John Langford, and Hanna
  Wallach.
\newblock A reductions approach to fair classification.
\newblock In {\em International Conference on Machine Learning}, pages 60--69.
  PMLR, 2018.

\bibitem{agarwal2019fair}
Alekh Agarwal, Miroslav Dud{\'\i}k, and Zhiwei~Steven Wu.
\newblock Fair regression: Quantitative definitions and reduction-based
  algorithms.
\newblock In {\em International Conference on Machine Learning}, pages
  120--129. PMLR, 2019.

\bibitem{celis2019classification}
L~Elisa Celis, Lingxiao Huang, Vijay Keswani, and Nisheeth~K Vishnoi.
\newblock Classification with fairness constraints: A meta-algorithm with
  provable guarantees.
\newblock In {\em Proceedings of the conference on fairness, accountability,
  and transparency}, pages 319--328, 2019.

\bibitem{kamishima2012fairness}
Toshihiro Kamishima, Shotaro Akaho, Hideki Asoh, and Jun Sakuma.
\newblock Fairness-aware classifier with prejudice remover regularizer.
\newblock In {\em Joint European conference on machine learning and knowledge
  discovery in databases}, pages 35--50. Springer, 2012.

\bibitem{kamiran2012decision}
Faisal Kamiran, Asim Karim, and Xiangliang Zhang.
\newblock Decision theory for discrimination-aware classification.
\newblock In {\em 2012 IEEE 12th International Conference on Data Mining},
  pages 924--929. IEEE, 2012.

\bibitem{pleiss2017fairness}
Geoff Pleiss, Manish Raghavan, Felix Wu, Jon Kleinberg, and Kilian~Q
  Weinberger.
\newblock On fairness and calibration.
\newblock {\em Advances in neural information processing systems}, 30, 2017.

\bibitem{Dua:2019}
Dheeru Dua and Casey Graff.
\newblock {UCI} machine learning repository, 2017.

\bibitem{bellamy2019ai}
Rachel~KE Bellamy, Kuntal Dey, Michael Hind, Samuel~C Hoffman, Stephanie Houde,
  Kalapriya Kannan, Pranay Lohia, Jacquelyn Martino, Sameep Mehta, Aleksandra
  Mojsilovi{\'c}, et~al.
\newblock Ai fairness 360: An extensible toolkit for detecting and mitigating
  algorithmic bias.
\newblock {\em IBM Journal of Research and Development}, 63(4/5):4--1, 2019.

\bibitem{lequy}
Tai Le~Quy, Arjun Roy, Vasileios Iosifidis, Wenbin Zhang, and Eirini Ntoutsi.
\newblock A survey on datasets for fairness-aware machine learning.
\newblock {\em WIREs Data Mining and Knowledge Discovery}, 12(3):e1452, 2022.

\bibitem{zhou2021improving}
Yan Zhou, Murat Kantarcioglu, and Chris Clifton.
\newblock Improving fairness of ai systems with lossless de-biasing.
\newblock {\em arXiv preprint arXiv:2105.04534}, 2021.

\bibitem{kamiran2009}
Faisal Kamiran and Toon Calders.
\newblock Classifying without discriminating.
\newblock In {\em 2009 2nd International Conference on Computer, Control and
  Communication}, pages 1--6, 2009.

\bibitem{VERBRAKEN2014505}
Thomas Verbraken, Cristián Bravo, Richard Weber, and Bart Baesens.
\newblock Development and application of consumer credit scoring models using
  profit-based classification measures.
\newblock {\em European Journal of Operational Research}, 238(2):505--513,
  2014.

\bibitem{calmon2017optimized}
Flavio Calmon, Dennis Wei, Bhanukiran Vinzamuri, Karthikeyan
  Natesan~Ramamurthy, and Kush~R Varshney.
\newblock Optimized pre-processing for discrimination prevention.
\newblock {\em Advances in neural information processing systems}, 30, 2017.

\bibitem{huang2019stable}
Lingxiao Huang and Nisheeth Vishnoi.
\newblock Stable and fair classification.
\newblock In {\em International Conference on Machine Learning}, pages
  2879--2890. PMLR, 2019.

\bibitem{friedler2019comparative}
Sorelle~A Friedler, Carlos Scheidegger, Suresh Venkatasubramanian, Sonam
  Choudhary, Evan~P Hamilton, and Derek Roth.
\newblock A comparative study of fairness-enhancing interventions in machine
  learning.
\newblock In {\em Proceedings of the conference on fairness, accountability,
  and transparency}, pages 329--338, 2019.

\bibitem{kamishima2013independence}
Toshihiro Kamishima, Shotaro Akaho, Hideki Asoh, and Jun Sakuma.
\newblock The independence of fairness-aware classifiers.
\newblock In {\em 2013 IEEE 13th International Conference on Data Mining
  Workshops}, pages 849--858. IEEE, 2013.

\bibitem{XIAO2021508}
Jin Xiao, Yadong Wang, Jing Chen, Ling Xie, and Jing Huang.
\newblock Impact of resampling methods and classification models on the
  imbalanced credit scoring problems.
\newblock {\em Information Sciences}, 569:508--526, 2021.

\bibitem{GUNNARSSON2021292}
Björn~Rafn Gunnarsson, Seppe {vanden Broucke}, Bart Baesens, María
  Óskarsdóttir, and Wilfried Lemahieu.
\newblock Deep learning for credit scoring: Do or don’t?
\newblock {\em European Journal of Operational Research}, 295(1):292--305,
  2021.

\bibitem{verbeke2022or}
Wouter Verbeke, Diego Olaya, Marie-Anne Guerry, and Jente Van~Belle.
\newblock To do or not to do? cost-sensitive causal classification with
  individual treatment effect estimates.
\newblock {\em European Journal of Operational Research}, 2022.

\bibitem{patterson2022}
David Patterson, Joseph Gonzalez, Urs Hölzle, Quoc Le, Chen Liang,
  Lluis-Miquel Munguia, Daniel Rothchild, David~R. So, Maud Texier, and Jeff
  Dean.
\newblock The carbon footprint of machine learning training will plateau, then
  shrink.
\newblock {\em Computer}, 55(7):18--28, 2022.

\end{thebibliography}

\begin{IEEEbiography}[{\includegraphics[width=1in,height=1.25in,clip,keepaspectratio]{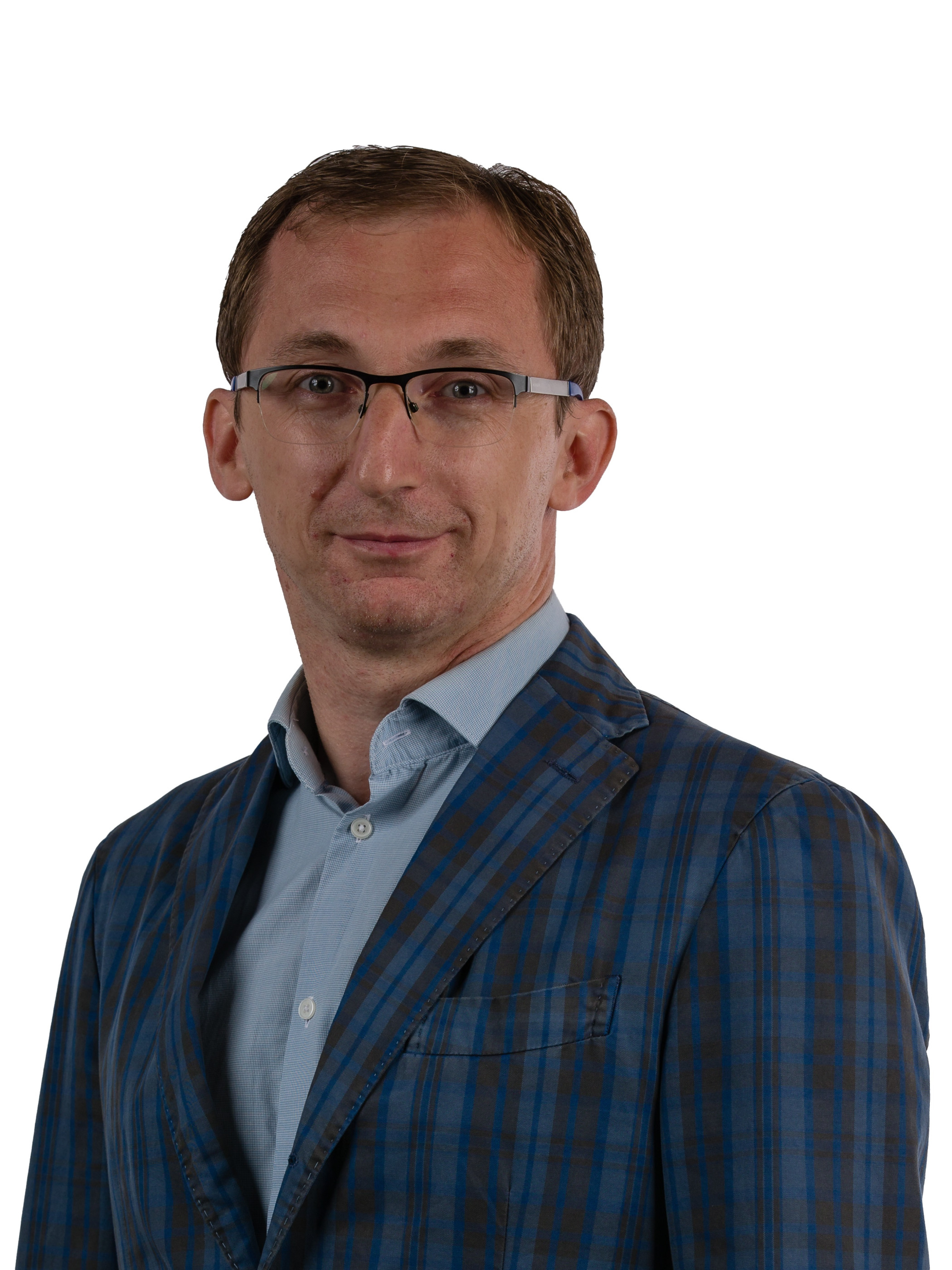}}]{Darie Moldovan}is an assistant professor of business analytics with Babe\c{s}
-Bolyai University (Romania). He obtained his PhD in business intelligence in 2012. His research is focused in the fields of analytics and operations research with applications in credit risk analysis, AI fairness, corporate governance and uplift modeling. He also worked in several industry projects related to credit risk prediction and consumer savings modeling. He teaches courses on Predictive Methods and Decision Support Systems for Business. Darie is a member of INFORMS and IEEE and servers as a reviewer specialized in analytics for several scientific journals.
\end{IEEEbiography}

\EOD

\end{document}